# Improving Warped Planar Object Detection Network For Automatic License Plate Recognition

Nguyen Dinh Tra, Nguyen Cong Tri, Phan Duy Hung

Computer Science Department, FPT University, Hanoi, Vietnam
`trandhe140661@fpt.edu.vn, trinche141519@fpt.edu.vn,`
`hungpd2@fe.edu.vn`

**Abstract:** This paper aims to improve the Warping Planer Object Detection Network (WPOD-Net) using feature engineering to increase accuracy. What problems are solved using the Warping Object Detection Network using feature engineering? More specifically, we think that it makes sense to add knowledge about edges in the image to enhance the information for determining the license plate contour of the original WPOD-Net model. The Sobel filter has been selected experimentally and acts as a Convolutional Neural Network layer, the edge information is combined with the old information of the original network to create the final embedding vector. The proposed model was compared with the original model on a set of data that we collected for evaluation. The results are evaluated through the Quadrilateral Intersection over Union value and demonstrate that the model has a significant improvement in performance.

## 1 Introduction

Convolutional Neural Network (CNN) is technically deep learning used effectively in feature extraction of images, widely use in Computer Vision to solve problems: Object Detection [1], Image segmentation [2], Recognition [3], Tracking [4] and Alignment [5] and has significant performance compared to using traditional machine learning. This article focuses on real-life license plate detection and recognition. One primary approach when it comes to domain transfer problems is using Warping Planer Object Detection Network [6].

The problem of license plate (LP) detection is not a new problem with many different methods, but when applying the CNN model, specifically YOLO [7] in the Object Detection problem, the accuracy increases compared to using machine learning. However, there is a limitation of this method that Object Detection only results in the area containing the license plate. If the license plate is at a non-front angle, the results can affect the extraction of license plate information.

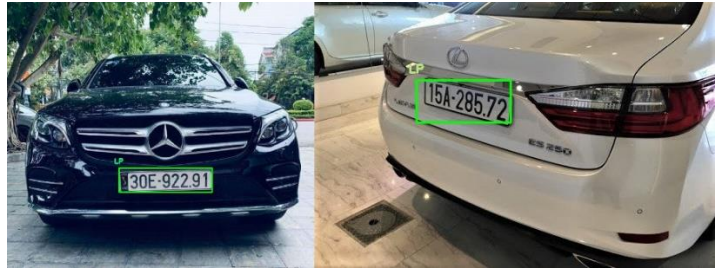

**Fig. 1**. License plate detection results using YOLO on image

In the Fig. 1, we can see that in the case of the front view (left image), the detected number plate area has almost no details of the body of the car. In the case of a front view and slightly tilted (right image), the detected license plate area includes quite a lot of excess image of the body.

In order to effectively detect objects such as license plates above when the input image is tilted too much, using WPOD-net is the first choice to detect the license plate area. This study gives an idea to improve that machine learning model architecture by adding information to the embedding vector. The edge information will be added as a CNN layer and the results will be compared with the original model on a set of data that we collected for evaluation.

## 2    Related Works

Sergio et al. designed WPOD-Net based on the idea of YOLO, SSD [8], STN (Spatial Transformer Network) [9]. As mentioned above YOLO and SSD only return 1 bounding box in the license plate regardless of the surrounding space. STN can detect non-rectangular regions, but it cannot handle multiple transformations at the same time, instead performing a single spatial transformation across the entire input. WPOD-Net return bounding box area surrounds the license plate and brings the number plate to the front view.

In the paper published in 2021: "A Flexible Approach for Automatic License Plate Recognition in Unconstrained Scenarios" [10], Sergio et al. present a method to improve WPOD model by add more convolutional layer in the end of model.

Vincent et al. in the paper "On Edge Detection" [11] present a commonly used digital image processing technique to find the contours of an image object. The author explained edge point: a pixel is considered an edge point if there is a rapid or sudden change in gray level (or color). Boundary: a set of consecutive boundary points. Edge detection is an image processing technique used to find the edges of objects in an image or can understand finding areas with a continuous loss in brightness (areas where there is a sharp difference in brightness).

In another paper published in 2009: "A Descriptive Algorithm for Sobel Image Detection" [12], Vincent et al. use a Gaussian filter to remove noise, smooth the image first to make the edge detection algorithm work better.

## 3   Methodology

This section presents the improved architecture, loss function and evaluation metric. The model is then used for license plate detection. Using image processing algorithms to find edges helps us to extract more information from the image. We can consider the Sobel filter as a special layer in the CNN model, more specifically the WPOD-Net that increases the accuracy of the License Plate detector.

### 3.1   The Architecture

The WPOD-net architecture is proposed using insights from [7-9]. The Fig. 2 shows how the WPOD-Net works. From input image through the forward process, we get output features maps include: 8 channels in which the first 2 channels are probability with/without license plate and remaining 6 channels are parameters to calculate the transformation matrix. To extract the license plate area, the authors consider a fixed size square part with white border around the cell in the output features map. If the probability of the cell's object is greater than threshold then the values of the cell's remaining 6 channels will calculate the transform matrix from square part to license plate area. The matrix can be used to bring the license plate to the front view. From this idea, the authors give the WPOD network architecture.

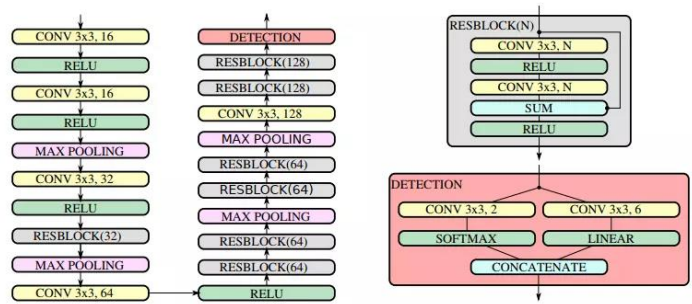

**Fig. 2**. WPOD Network Architecture

From the original model, with the idea that adding edge information to the embedding vector can potentially increase the performance of the model, we have proposed a new architecture as shown in the Fig. 3.

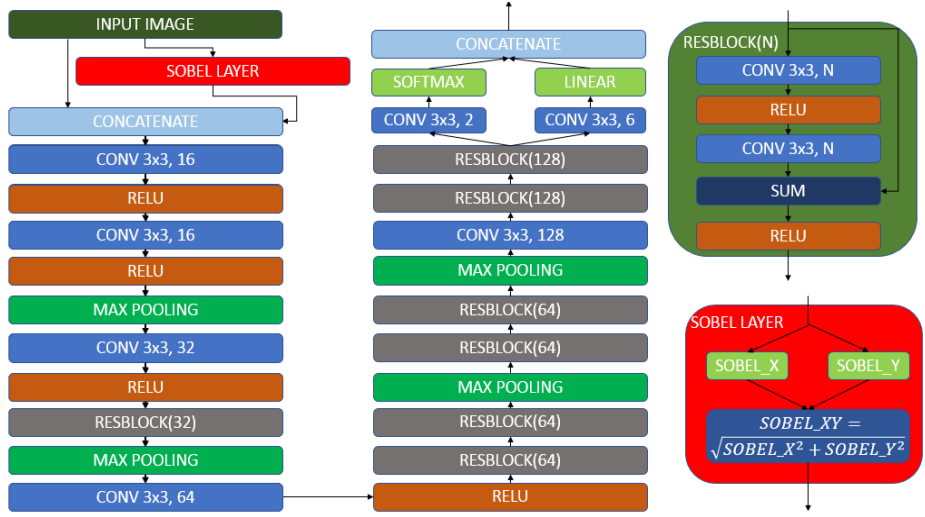

**Fig. 3**. The proposed model

The Sobel filter is based on convolving an image with a set of pre-determined 3x3 kernels. It is an example of a gradient-based edge detection algorithm. The first kernel computes the horizontal gradient at each pixel (SOBEL_X in the Fig. 3), while the second kernel computes the vertical gradient (SOBEL_Y in the Fig. 3).

These kernels can be used to compute the horizontal and vertical gradients at each pixel, and the resulting gradients can then be combined to determine the edge strength at each pixel (SOBEL_XY in the Fig. 3).

### 3.2  Loss Function

In this work, the mean-squared-error (MSE) loss is used to estimate the error between a warped version of the canonical square and the normalized annotated points of the LP. The binary-cross-entropy (BCS) loss is used to handle the probability of having/not having an object at each pixel in final feature map. For an input image with height $H$ and width $W$, and network stride given by $N_s = 2^4$ (four max pooling layers), the network output feature map consists of an $M \times N \times 8$ volume, where $M = H/N_s$ and $N = W/N_s$. For each point cell $(m, n)$ in the feature map, there are eight values to be estimated: the first two values ($v_1$ and $v_2$) are the object/non-object probabilities, and the last six values ($v_3$ to $v_8$) are used to build the local affine transformation $T_{mn}$ given by:

$$T_{mn}(q) = \begin{bmatrix} \max(v_3, 0) & v_4 \\ v_5 & \max(v_6, 0) \end{bmatrix} q + \begin{bmatrix} v_7 \\ v_8 \end{bmatrix},$$

where the max function used for $v_3$ and $v_6$ was adopted to ensure that the diagonal is positive (avoiding undesired mirroring or excessive rotations).

To match the network output resolution, the points $\boldsymbol{p}_i$ are re-scaled by the inverse of the network stride, and re-centered according to each point $(m, n)$ in the feature map. This is accomplished by applying a normalization function

$$A_{mn}(\boldsymbol{p}) = \frac{1}{\alpha}\left(\frac{1}{N_s}\boldsymbol{p} - \begin{bmatrix} n \\ m \end{bmatrix}\right),$$

where $\alpha$ is a scaling constant that represents the side of the fictional square. We set $\alpha = 7.75$, which is the mean point between the maximum and minimum LP dimensions in the augmented training data divided by the network stride.

With $T_{mn}(\boldsymbol{q}_i)$ and $A_{mn}(\boldsymbol{p}_i)$ are defined by Sergio et al. in WPOD net, location loss as follow:

$$f_{location}(m, n) = \sum_{i=1}^{4} (T_{mn}(\boldsymbol{q}_i) - A_{mn}(\boldsymbol{p}_i))^2$$

Confidence loss as follow:

$$f_{\text{confidence}}(m, n) = \log \text{loss}(\mathbb{I}_{obj}, v_1) + \log \text{loss}(1 - \mathbb{I}_{obj}, v_2)$$

Total loss function is given by a combination of location loss and confidence loss:

$$\text{Total\_loss} = \sum_{m=1}^{M} \sum_{n=1}^{N} \left[\mathbb{I}_{obj} f_{location}(m, n) + f_{\text{confidence}}(m, n)\right]$$

where $\mathbb{I}_{obj}$ is the object indicator function that returns 1 if there is an object at point (m, n) or 0 otherwise.

### 3.3 Evaluation metric

The Quadrilateral Intersection over Union (qIoU) is used to evaluate the accuracy of the detection model.

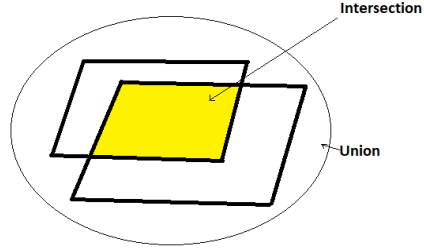

**Fig. 4**. Quadrilateral Intersection over Union (qIoU)

In the Fig. 4, we can see how to calculate qIoU, higher qIoU proves that the model is more effective for detecting license plates.

## 4 Implementation & Evaluation

### 4.1 Implement

This CNN model is implemented in Tensorflow [13]. Whole experiments were performed on an NVIDIA GeForce RTX 3090 GPU.

The model is trained by using a per-batch training strategy and use Adam's algorithm with learning rate 10-3, batch size 64 for 300,000 iterations.

In each batch, we randomly use augment data. The following augmentation transforms are used:

- Rectification
- Aspect-ratio
- Centering
- Scaling
- Rotation
- Mirroring
- Translation
- Cropping
- Colorspace
- Annotation

### 4.2 Dataset

We create a training dataset of 363 images, including some images taken from the AOLP dataset [14], some data of Vietnamese car and motorbike license plates we collected from Internet. Specifically we have: 50 images from the AOLP dataset, 105 images of cars, and 208 images of motorbikes. An independently generated test data, 175 images include: 25 images from the AOLP dataset, 50 images of cars and 100 images of motorbikes.

For each image, we manually annotated the 4 corners of the LP in the picture. A few samples are shown in the Fig. 4.

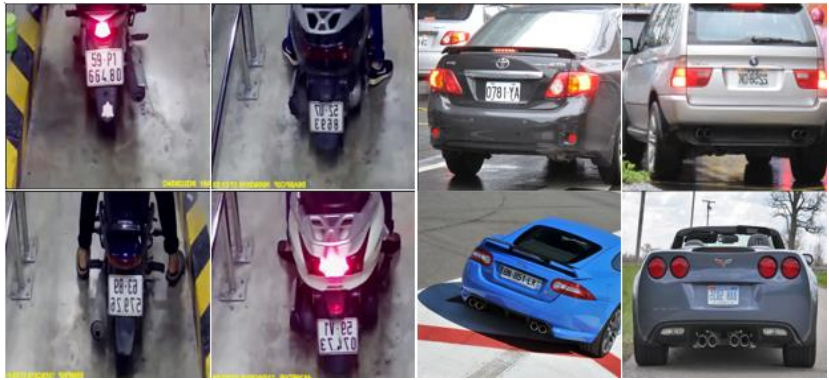

**Fig. 4**. Examples of the LPs in the training dataset

### 4.3 Comparison to Other Methods

Model performance is compared to WPOD-net and iWPOD-net [10].

**Table 1**. Quantitative evaluation

| Method | qIoU |
| --- | --- |
| WPOD-net | 84.81% |
| iWPOD-net | 84.32% |
| Proposed Model | *85.81%* |

WPOD-net, iWPOD-net and the proposed model are trained by the same dataset, loss function, optimization algorithm, hyper parameters. The results in Table 1 show that the improved model gives significantly better results in qIoU (from 1 to 1.5%) than the original model when evaluated on our dataset.

## 5 Conclusion & Future Works

This paper proposes a method to improve the performance of WPOD network by adding edge knowledge to the embedding vector of the network. The results of the article show that when applied to the license plate detection problem, the quality of license plate detection is significantly improved.

Accurate object detection, especially in cases where the subject image is tilted, will help better in the subsequent information extraction. The paper can be a good reference for many machine learning and data mining problems.